\DocumentMetadata{%
	pdfstandard=A-3u,
	pdfversion=1.7,
	lang=en-US,
}

\documentclass[balance,colorlinks,upint,subscriptcorrection,varvw,mathalfa=cal=boondoxo]{asmeconf}

\usepackage{algorithm}
\usepackage{algpseudocode}
\hypersetup{%
	pdfauthor={Nicolas Gautier},									  
	pdftitle={Optimizing Robot Positioning against on Placement Inaccuracies: A Study on the Fanuc CRX10iA/L},                  
	pdfkeywords={Welding & Cutting applications, serial robot, trajectory planning, simulation},
	pdfsubject = {Optimizing Robot Positioning},			  
}

\begin{document}


\ConfName{}
\ConfAcronym{I}
\ConfDate{}
\ConfCity{}
\PaperNo{ }


\title{Optimizing Robot Positioning against Placement Inaccuracies: \\
A Study on the Fanuc CRX10iA/L} 
 
%
%
%

\SetAuthors{%
	Nicolas Gautier \affil{1}\affil{2}, 
        Yves Guillermit \affil{1},
	Mathieu Porez \affil{2},
    David Lemoine \affil{2},
	Damien Chablat \affil{2}\CorrespondingAuthor{Damien.Chablat@cnrs.fr}
	}

\SetAffiliation{1}{Weez-U Welding, 101 rue de Coulmiers, 44 000 Nantes, France}
\SetAffiliation{2}{Nantes Université, École Centrale Nantes, IMT Atlantique, CNRS, LS2N, UMR 6004, F-44000 Nantes, France}


\maketitle

\versionfootnote{Documentation for \texttt{asmeconf.cls}: Version~\versionno, \today.}


\begin{abstract}
This study presents a methodology for determining the optimal base placement of a Fanuc CRX10iA/L collaborative robot for a desired trajectory corresponding to an industrial task. The proposed method uses a particle swarm optimization algorithm that explores the search space to find positions for performing the trajectory. An $\alpha$-shape algorithm is then used to draw the borders of the feasibility areas, and the largest circle inscribed is calculated from the Voronoi diagrams. The aim of this approach is to provide a robustness criterion in the context of robot placement inaccuracies that may be encountered, for example, if the robot is placed on a mobile base when the system is deployed by an operator. The approach developed uses an inverse kinematics model to evaluate all initial configurations, then moves the robot end-effector along the reference trajectory using the Jacobian matrix and assigns a score to the attempt. For the Fanuc CRX10iA/L robot, there can be up to 16 solutions to the inverse kinematics model. The calculation of these solutions is not trivial and requires a specific study that planning tools such as MoveIt cannot fully take into account. Additionally, the optimization process must consider constraints such as joint limits, singularities, and workspace limitations to ensure feasible and efficient trajectory execution.
\end{abstract}

\section{Introduction}
Optimization of a robot’s placement for complex, sometimes one-off, industrial tasks such as welding or cutting represents a major challenge \cite{gautier2024comparison}, particularly when the workpieces are large relative to the robot and the trajectory to be followed is highly constrained in terms of orientation. For example, a trajectory resulting from the intersection of two cylinders imposes strict requirements on both position and orientation, making it difficult to determine an optimal placement that ensures feasible movements. These constraints are even more critical as positioning inaccuracies can affect execution robustness, e.g. when the robot is mounted on a mobile platform. This issue remains a major challenge, and robot placement still relies largely on iterative testing and human intuition.


In the literature, several approaches have been proposed to address this problem. The approach presented in \cite{spensieri2016optimal} is based on optimizing the base position of an industrial robot to minimize cycle time while ensuring the absence of collisions with the environment. Similarly, \cite{zbiss2024automatic} focuses on optimizing the number of robots and their positioning to maximize paint coverage while reducing the risk of collisions between robots. Other studies have focused on minimizing the robot's dynamic effects. In \cite{stradovnik2024workpiece}, the authors propose to place the workpiece to minimize joint efforts, thereby improving task quality. Likewise, \cite{vosniakos2010improving} aims to reduce joint torques in a milling application due to the low rigidity and high flexibility of robots compared to traditional CNC machines. For mobile manipulation tasks, \cite{zhang2023base} proposes a multi-objective optimization approach to improve coverage and the efficiency of the robot’s placement on its mobile base. Finally, \cite{saini2024planning} develops a strategy that combines planning and heuristics to optimize robot placement in object-grasping tasks, taking into account robot reach and surrounding obstacles.

In this paper, we propose a new approach that combines kinematic trajectory simulation (including Inverse Kinematics Model (IKM), motion constraints, singularities, joint limits, and robot self-collision detection), workspace exploration using a Particle Swarm Optimization (PSO) algorithm, and the evaluation of the largest feasibility area. Our method stands out by introducing a robustness criterion to compensate for positioning inaccuracies, ensuring better adaptability to real execution conditions. Moreover, its computational efficiency allows quick integration of a robot into industrial environments requiring regular adjustments.

In the next two sections, we detail our methodology and the different steps of our approach. We then present the experimental results obtained before concluding with perspectives for improvement and future applications.
\section{Simulation of robotic trajectories}
To model, simulate and plan a robotic task, the robot's trajectories in space are defined by a set of end-effector poses. Each pose includes the vector position ${\bf P}= [x, y, z]$ and the set of Euler angles $[\varphi, \theta, \psi]$ (following the \lq Z-Y-X' convention) in the world frame. The pose $p_i$ is represented by the homogeneous transformation matrix ${}^W {\bf T}_E^i$, as follow \cite{khalil2002modeling}
\begin{equation}
        {}^W\textbf{T}_{E}^{(p_i)} = \begin{bmatrix}
            \textbf{R}^{(p_i)} & \textbf{P}^{(p_i)} \\
            0 & 1 
        \end{bmatrix}\text{ ,} 
\end{equation}
with
\begin{eqnarray}\label{eq:wte}     
        {\bf R}  &=& Rot(z, \psi) \cdot Rot(y, \theta) \cdot Rot(x, \varphi) \\
           &=&
        \begin{bmatrix}
            C_\theta C_\psi & S_{\varphi} S_{\theta} C_\psi - C_{\varphi} S_{\psi} & C_{\varphi} S_{\theta} C_\psi + S_{\varphi} S_{\psi}\\
            C_\theta S_{\psi} & S_{\varphi} S_{\theta} S_{\psi} + C_{\varphi} C_\psi & C_{\varphi} S_{\theta} S_{\psi} - S_{\varphi} C_\psi \\
            -S_{\theta} & S_{\varphi} C_\theta & C_{\varphi} C_\theta\\
        \end{bmatrix} \text{ ,} \nonumber
\end{eqnarray}
%
where, \(C_\alpha\) denotes the cosine of \(\alpha\), and \(S_\alpha\) denotes the sine of \(\alpha\). For a pose of the robot base ${}^W \textbf{T}_0$ (which is the parameter to optimize) and a configuration of the Terminal Center Point (TCP or end-effector) ${}^6 {\bf T}_E$, each point of the trajectory can be expressed in the robot's base frame by the following relation: 
\begin{equation}
        {}^0\textbf{T}_6^{(p_i)} = {}^W \textbf{T}_0^{-1}\cdot{}^W \textbf{T}_E^{(p_i)}\cdot{}^6 \textbf{T}_E^{-1} \text{ .}
\end{equation}
In the paper, we will use the Fanuc CRX10iA/L robot as an example. 
The modified Denavit Hartenberg (DH) parameters of this robot (Khalil-Kleinfinger convention, see \cite{khalil1986new}) are given in Table \ref{tab:DH_parameters}. 
\begin{table}[!h]
    \centering
    \begin{tabular}{|c|c|c|c|c|}
    \hline
    $i$ & $d_i$ (m) & $a_i$ (m) & $\alpha_i$ (rad) & $\theta_i$ (rad)\\
    \hline
    1 & 0.245 & 0 & 0 & $q_1$\\
    2 & 0 & 0 & $\pi/2$ & $q_2$ \\
    3 & 0 & 0.710 & 0 & $q_3$ \\
    4 & 0.540 & 0 & $-\pi/2$ & $q_4$ \\
    5 & 0.150 & 0 & $\pi/2$ & $q_5$\\
    6 & 0.160 & 0 & $-\pi/2$  & $q_6$\\
    \hline
    \end{tabular}
    \caption{Modified DH parameters of the Fanuc CRX10iA/L robot.}
    \label{tab:DH_parameters}
\end{table}

In the rest of the document the position of joint $i$ will be noted $q_i$. We will fix the values of the twist angles $\alpha$, while the values of the joint-to-joint distance $d$ and the offset $a$  will remain symbolic. The developed equations will thus be valid for all robots with a similar kinematic architecture. It is worth noting that 6R anthropomorphic robotic arms with a spherical wrist are a special case of this architecture, where $d_5=d_6=0$. For numerical applications, we will use the length values from the Fanuc CRX10iA/L robot.
%
The joint limits of the Fanuc CRX10iA/L robot are given in Table \ref{tab:limit_articulaire} for this set of DH parameters.
\begin{table}[H]
    \centering
    \begin{tabular}{|c|c|c|c|c|c|c|}
    \hline
    i & 1 & 2 & 3 & 4 & 5 & 6 \\
    \hline
    min ($\rm rad$) & -$\pi$ & -$\pi/2$ & -$\pi$ & -$19\pi/18$ & -$\pi$ & -$5\pi/4$\\
    max ($\rm rad$)& $\pi$ & $3\pi/2$ & $2\pi$ & $19\pi/18$ & $\pi$ & $5\pi/4$\\
    \hline
    \end{tabular}
    \caption{Joint limits of the Fanuc CRX10iA/L robot}
    \label{tab:limit_articulaire}
\end{table}
%
%
 Finally, to calculate the position of the end-effector according to its base, we use transformation matrices between each link. Between two bodies $i$ and $j$, this matrix is given by:

\begin{equation}
    {}^i\textbf{T}_j = \begin{bmatrix}
        C_{\theta_j } & -S_{\theta_j} & 0 & a_j\\
        C_{\alpha_j} S_{\theta_j} & C_{\alpha_j} C_{\theta_j} & -S_{\alpha_j} & -d_jS_{\alpha_j}
        \\
        S_{\alpha_j} S_{\theta_j} & S_{\alpha_j} C_{\theta_j} & C_{\alpha_j} & d_jC_{\alpha_j} \\
        0 & 0 & 0 &1
\end{bmatrix}\text{ .}
\end{equation}

\subsection{Computation of inverse kinematic model}
To create a continuous trajectory, we must avoid singularities, joint limits, and self-collision. However, these factors depend on the robot's initial position, and different initial positions can lead to different outcomes. Therefore, it is essential to evaluate all possible starting postures, i.e., all solutions to the inverse kinematics problem for the starting point of the trajectory. Inverse kinematics is a fundamental problem in robotics, which involves calculating the joint angles necessary for a robot to reach a given position and orientation of its end-effector. This problem is often complex, especially for robots with a cuspidal morphology \cite{wenger2023review}, such as the Fanuc CRX10iA/L, which can have up to 16 distinct solutions, and for which no general analytical solution exists \cite{salunkhe2024kinematic}.
%
To address this problem, several methods have been proposed. One of them relies on the use of iterative algorithms that exploit higher-order derivatives to converge towards a solution \cite{lloyd2022fast}. Other approaches include polynomial resolution, such as the one applied to the Kinova Gen3 Lite manipulator, where a univariate polynomial equation in 
$q_1$ (i.e. the first joint position of the robot) is solved, with the other variables determined by backward substitution \cite{zohour2021kinova}.
%
We used a method similar to the one employed for an anthropomorphic robot. According Paul's method, position and orientation are decomposed to separate the problem into independent solutions for $q_1, q_2$, and $q_3$, and then for $q_4, q_5$, and $q_6$ \cite{khalil2002modeling}. However, the wrist offset at joint 5 makes the position of the wrist dependent on $q_4$. To address this, we discretize $q_4$ over the interval $[-\pi, \pi]$ in order to compute $q_1, q_2$, and $q_3$. From a final equation derived from the rotation matrix, we can compute a residual that indicates a pair of solutions when it vanishes.
%
This method has the advantage of obtaining the set of solutions to the inverse kinematics problem, provided that the discretization step of $q_4$ is fine enough. It allows solving 4 equations instead of a single one of degree 16. This is particularly useful when using non-mathematical programming languages such as Python or C.

\begin{figure}[H]
    \centering
    \includegraphics[width=1\linewidth]{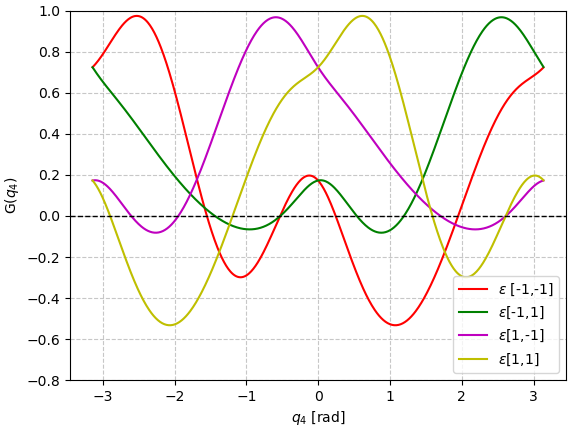}
    \caption{Four residual curves corresponding to the four couples $[\epsilon_1, \epsilon_2]$ for the pose $[-0.2406, -0.1188, 0.5603, 2.6204, 1.1236, 0.4276]$ as a function of $q_4$}.
    \label{fig:courbe_q4}
\end{figure}

\begin{figure}[!ht]
	\centering
	\begin{subfigure}{0.155\textwidth}
		\includegraphics[width=\textwidth]{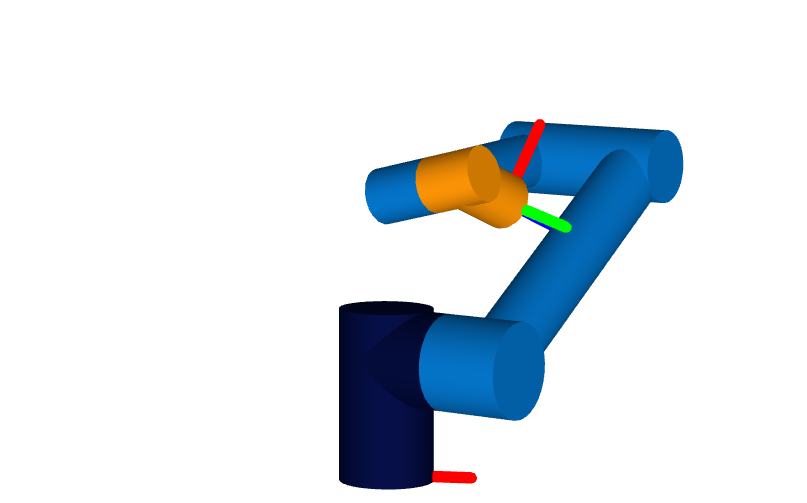}
		\caption{Posture 1}
	\end{subfigure}
	\begin{subfigure}{0.155\textwidth}
		\includegraphics[width=\textwidth]{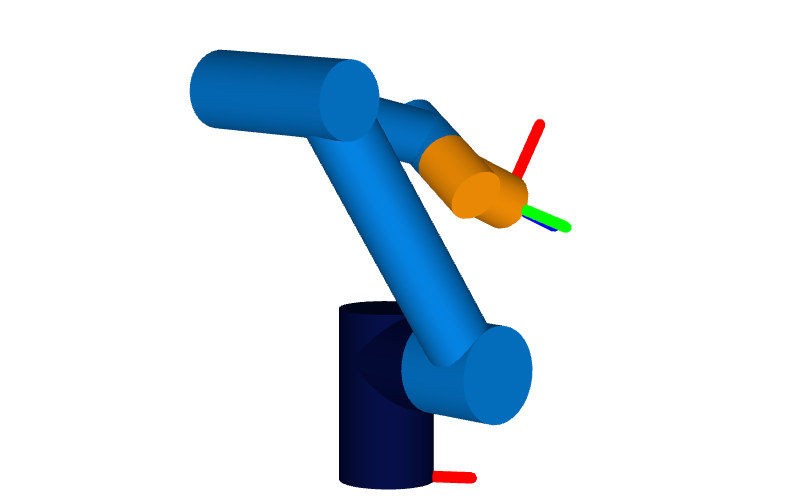}
		\caption{Posture 2}
	\end{subfigure}
	\begin{subfigure}{0.155\textwidth}
		\includegraphics[width=\textwidth]{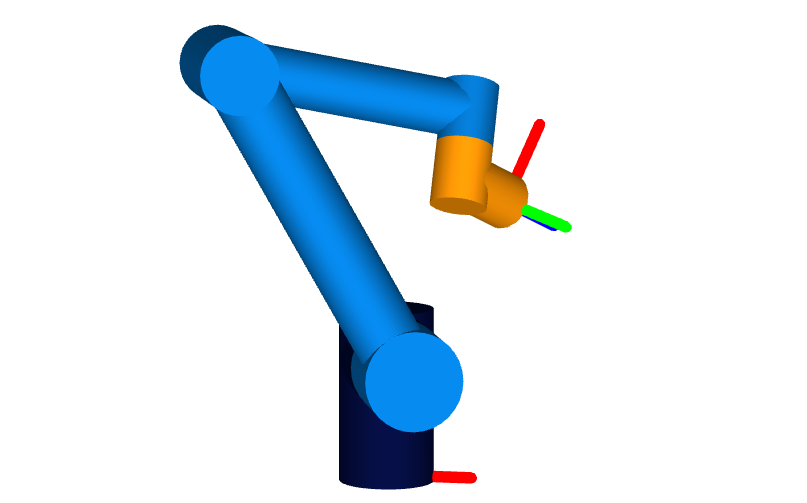}
		\caption{Posture 3}
	\end{subfigure}
	\begin{subfigure}{0.155\textwidth}
		\includegraphics[width=\textwidth]{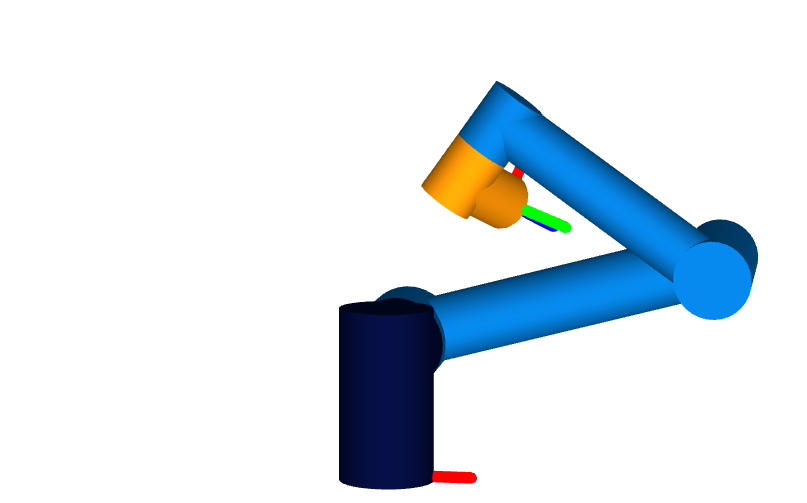}
		\caption{Posture 4}
	\end{subfigure}
	\begin{subfigure}{0.155\textwidth}
		\includegraphics[width=\textwidth]{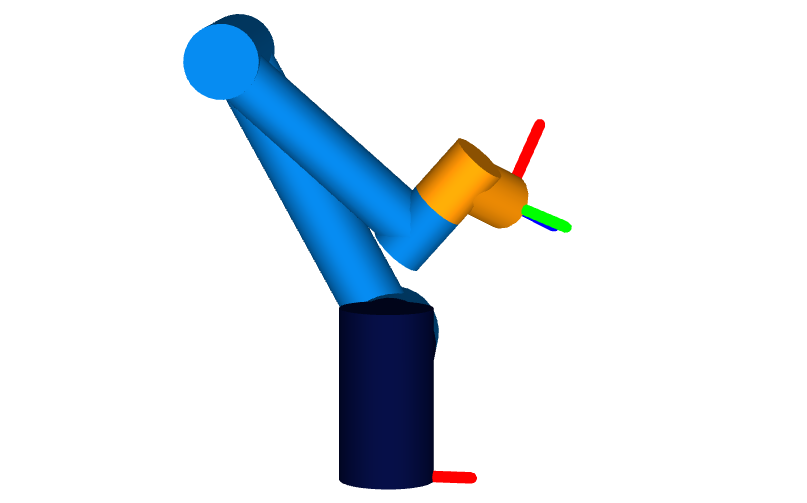}
		\caption{Posture 5}
	\end{subfigure}
	\begin{subfigure}{0.155\textwidth}
		\includegraphics[width=\textwidth]{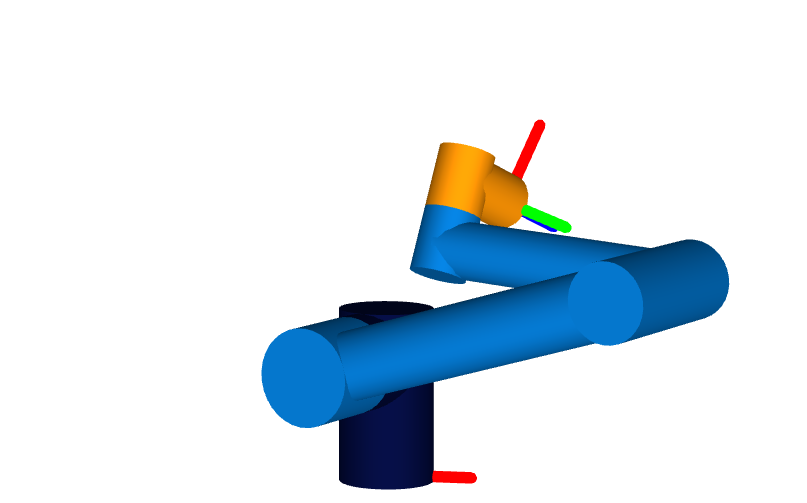}
		\caption{Posture 6}
	\end{subfigure}
        \begin{subfigure}{0.155\textwidth}
		\includegraphics[width=\textwidth]{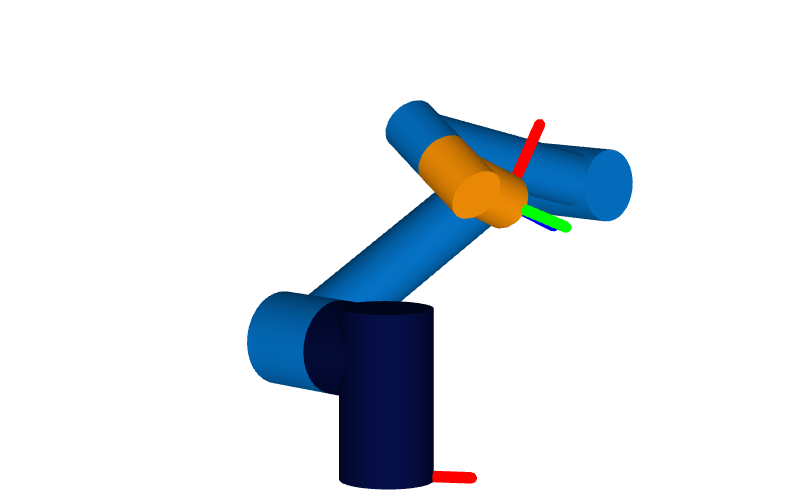}
		\caption{Posture 7}
	\end{subfigure}
	\begin{subfigure}{0.155\textwidth}
		\includegraphics[width=\textwidth]{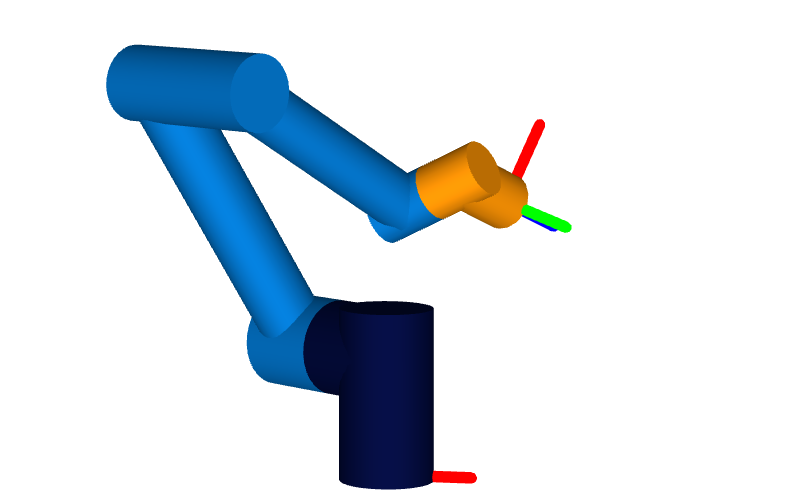}
		\caption{Posture 8}
	\end{subfigure}
	\begin{subfigure}{0.155\textwidth}
		\includegraphics[width=\textwidth]{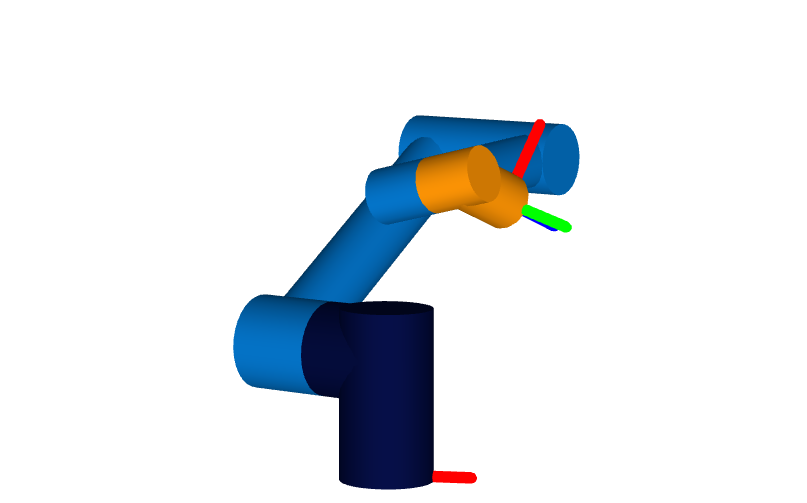}
		\caption{Posture 9}
	\end{subfigure}
	\begin{subfigure}{0.155\textwidth}
		\includegraphics[width=\textwidth]{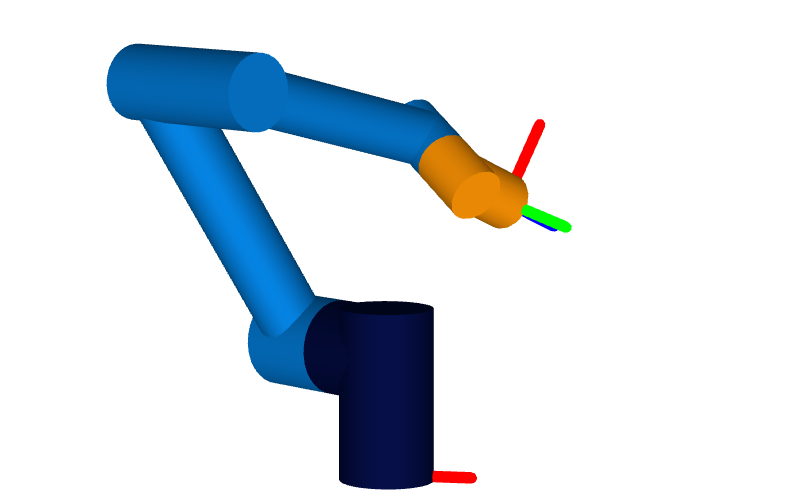}
		\caption{Posture 10}
	\end{subfigure}
	\begin{subfigure}{0.155\textwidth}
		\includegraphics[width=\textwidth]{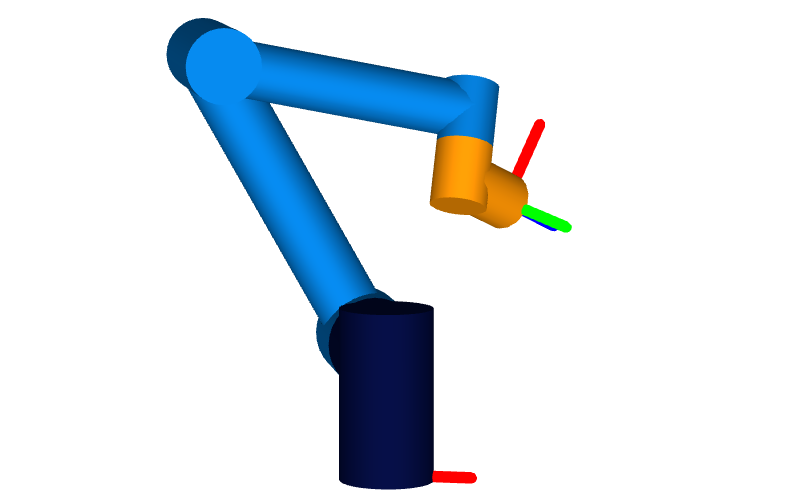}
		\caption{Posture 11}
	\end{subfigure}
	\begin{subfigure}{0.155\textwidth}
		\includegraphics[width=\textwidth]{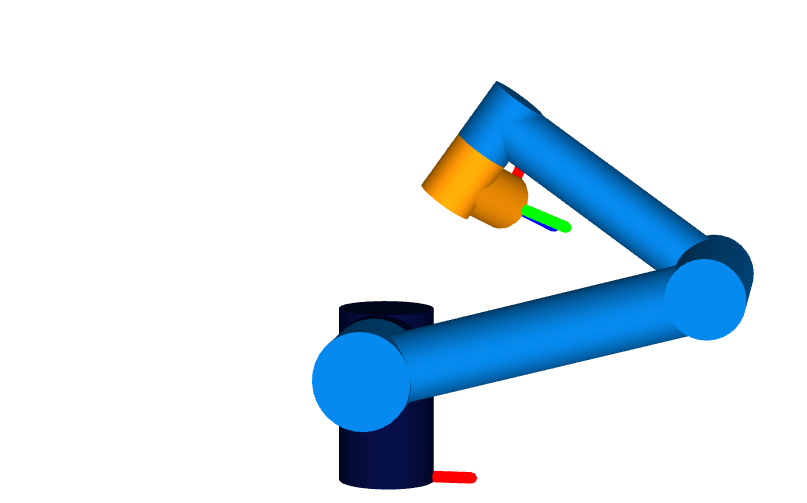}
		\caption{Posture 12}
	\end{subfigure}
        \begin{subfigure}{0.155\textwidth}
		\includegraphics[width=\textwidth]{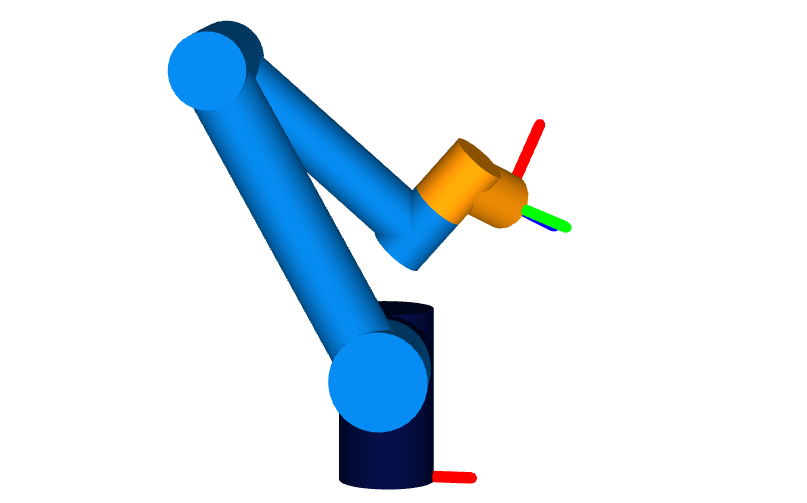}
		\caption{Posture 13}
	\end{subfigure}
	\begin{subfigure}{0.155\textwidth}
		\includegraphics[width=\textwidth]{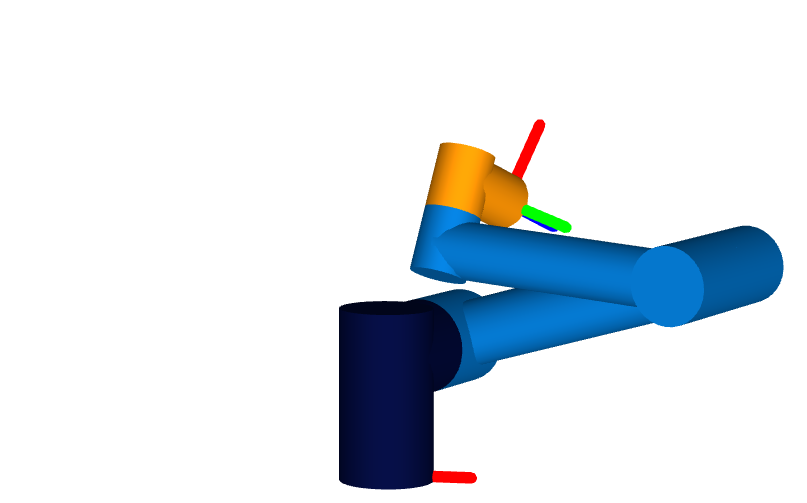}
		\caption{Posture 14}
	\end{subfigure}
	\begin{subfigure}{0.155\textwidth}
		\includegraphics[width=\textwidth]{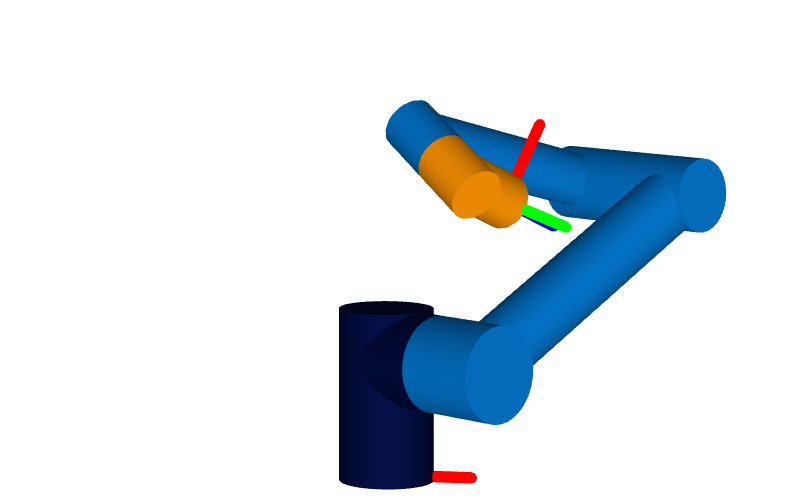}
		\caption{Posture 15}
	\end{subfigure}
	\begin{subfigure}{0.155\textwidth}
		\includegraphics[width=\textwidth]{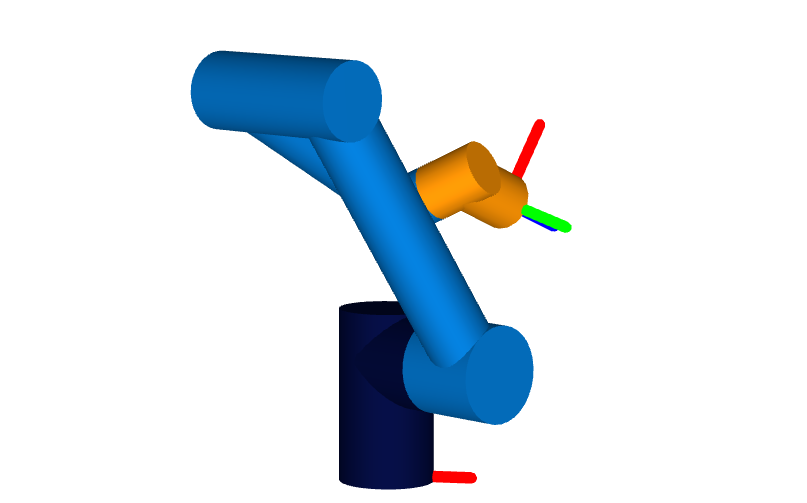}
		\caption{Posture 16}
	\end{subfigure}
	\caption{Visualization of the 16 inverse kinematic solutions for the pose
    $[-0.2406,  -0.1188,  0.5603,  2.6204,  1.1236,  0.4276]$.}
	\label{fig:post_ex_iks}
\end{figure}

%
The position equations of ${}^0{\bf T}_5$, centre of the wrist, are given by:
\begin{equation}\label{eq:0T5_pos}
    \begin{cases} P_{x} = x - d_6 r_{13} \\ P_y = y - d_6r_{23} \\ P_z = z - d_6r_{33}\end{cases}\text{ ,}
\end{equation}
%
where $r_{ij}$ is the component in row $i$ and column $j$ of the rotation matrix $\textbf{R}$. By pre-multiplying (\ref{eq:0T5_pos}) by ${}^1 {\bf T}_0$, the position equations (\ref{eq:0T5_pos}) become:
\begin{equation} \label{eq:1T5_pos}
    \begin{cases}C_1P_x + S_1P_y = a_3C_2-d_4S_{23}+d_5C_{23}S_4 \\ -S_1P_x + C_1P_y = -d_5C_4 \\ P_z - d_1 = d_4C_{23} + a_3S_2 + d_5S_{23}S_4 \end{cases}\text{ .}
\end{equation}
The second equation of (\ref{eq:1T5_pos}) is of the form \( XS_1 + YC_1 = Z \), where \(X,Y\) and \(Z\) are generalized variables. The solutions to solve for \( q_1 \) are given by:
\begin{equation}
q_1 = \rm{atan2}(S_1, C_1) \text{ ,}
\end{equation}
where:
\begin{eqnarray*}
S_1 &=& \dfrac{XZ+\epsilon_1 Y \sqrt{X^2+Y^2-Z^2}}{X^2+Y^2} \text{ , and}\\
C_1 &=& \dfrac{YZ-\epsilon_1 X \sqrt{X^2+Y^2-Z^2}}{X^2+Y^2} \text{ ,}
\end{eqnarray*}
%
with \( \epsilon_1 \in \{-1, 1\} \). Since \( q_1 \) is known, we can pre-multiply our matrix ${}^1{\bf T}_5$ by ${}^2{\bf T}_1$. The position equations become:
\begin{equation*}
        \begin{cases}(P_z-d_1)S_2 + (P_xC_1 + P_yS_1)C_2 = a_3 -d_4S_3 + d_5S_4C_3 \\ (P_z-d_1)C_2 - (P_xC_1 + P_yS_1)S_2 = d_4C_3 + d_5S_4S_3 \\ P_xS_1 - P_yC_1 = d_5C_4 \end{cases}\text{ ,}
\end{equation*}
%
which can be written in the form:
\begin{equation*}
        \begin{cases}W_1C2 + W_2S_2 = XC_3 + YS_3+Z_1 \\ W_1S2 - W_2C_2 = -XS_3 + YC_3 \end{cases}\text{ .}
\end{equation*}
%
By squaring and adding the two equations, we are led to solve an equation of the form $B_1 S_3 + B_2 C_3 = B_3$ with $B_1 = 2Z_1 Y$, $B_2 = 2Z_1 X$, and $B_3 = W_1^2 + W_2^2 - X^2 - Y^2 - Z_1^2$. The solutions for $q_3$ are given by:
\begin{equation}\label{eqn:P25}
q_3 = \rm{atan2}(S_3, C_3)\text{ ,}
\end{equation}
where:
\begin{eqnarray*}
        S_3 &=& \dfrac{B_1B_3+\epsilon_2 B_2 \sqrt{B_1^2+B_2^2-B_3^2}}{B_1^2+B_2^2} \text{ , and}\\ C_3 &=& \dfrac{B_2B_3-\epsilon_2 B_1 \sqrt{B_1^2+B_2^2-B_3^2}}{B_1^2+B_2^2}\text{ ,}
\end{eqnarray*}
%
%
with $\epsilon_2 \in \{-1, 1\}$. Knowing $q_3$, the two equations previously used in (\ref{eqn:P25}) become of the form:
\begin{equation}
    \begin{cases}X_1S_2 + Y_1C_2 = Z_1 \\ X_2S_2 + Y_2C_2 = Z_2 \end{cases}\text{ .}
\end{equation}
%
The solution for $q_2$ is:
\begin{equation}
    q_2= \rm{atan2}(S_2, C_2)\text{ ,}
\end{equation}
where:
\begin{equation*}
     S_2 = \dfrac{Z_1Y_2 - Z_2Y_1}{X_1Y_2-X_2Y_1} \text{ , and } C_2 = \dfrac{Z_2X_1-Z_1X_2}{X_1Y_2-X_2Y_1}\text{ .}
\end{equation*}
%
To obtain a new equation to calculate the residual of $q_4$, we use the orientation equations of the rotation matrix ${}^3{\bf R}_6$. The $r_{13}$ and $r_{33}$ components of this matrix give us the following equations:
\begin{equation}\label{eq:3R6_rot}
    \begin{cases} -C_4S_5 = r_{33}S_{23}+r_{13}C_1C_{23} + r_{23}S_1C_{23}\\ S_4S_5 = r_{13}S_1-r_{23}C_1\end{cases}\text{ .}
\end{equation}
%
By multiplying the equations of (\ref{eq:3R6_rot}) by $S_4$ and $C_4$ respectively and adding them, we obtain the following equation, named residual equation and denoted by $G$:
\begin{eqnarray}\label{eq:residual_eq}
        G(q_4, \epsilon) &=& S_4(r_{33}S_{23}+r_{13}C_1C_{23} + r_{23}S_1C_{23}) \\&+& C_4(r_{13}S_1-r_{23}C_1) \text{ .}\nonumber
\end{eqnarray}
%
According to the four possible combinations of sign coefficients $[\epsilon_1, \epsilon_2]$, from (\ref{eq:residual_eq}), we obtain four equations. When the residual equation is zero, the quadruple $\{q_1, q_2, q_3, q_4\}$ constitutes a solution of the inverse kinematic. Each equation has at most four solutions. Figure \ref{fig:courbe_q4} illustrates an example of residual evolutions as a function of $q_4$. This example is given for the pose $[x, y, z, \varphi, \theta, \psi]$ = $[-0.2406, -0.1188, 0.5603, 2.6204, 1.1236, 0.4276]$ where 16 real solutions exist, as each of the four equations has four real solutions.
%
%
To obtain the last four equations to calculate $q_5$ and $q_6$, we use the orientation equations of the rotation matrix ${}^4{\bf R}_6$. The $r_{13}$ and $r_{33}$ components determine $q_5$:
\begin{equation}
    q_5 = \rm{atan2}(S_5, C_5)\text{ ,}
\end{equation}
where:
\begin{eqnarray*}
S_5 &=& S_4(r_{13}S_1 -r_{23}C_1)  -C_4(r_{33}S_{23} + C_{23}(r_{13}C_1+r_{23}S_1)) \text{ ,}\\ 
C_5 &=&  r_{33}C_{23} - S_{23}(r_{13}C_1 + r_{23}S_1)\text{ ,}
\end{eqnarray*}
and the $r_{21}$ and $r_{22}$ components determine $q_6$:
\begin{equation} 
q_6 = \rm{atan2}(S_6, C_6)\text{ ,}
\end{equation}
where:
\begin{eqnarray*} 
S_6 &=& C_4(r_{21}C_1-r_{11}S_1) - r_{31}S_{23}S_4 \\&-& C_{23}S_4(r_{11}C_1+r_{21}S_1)\text{ , and}\\
C_6 &=& C_4(r_{22}C_1-r_{12}S_1) - r_{32}S_{23}S_4 \\&-& C_{23}S_4(r_{12}C_1+r_{22}S_1)\text{ .}
\end{eqnarray*}
%
%
The solutions for the previous example are given in Table \ref{tab:sol_iks}, and Figure \ref{fig:post_ex_iks} illustrates the corresponding robot postures.
%

\begin{table}[!ht]
	\centering
	\begin{tabular}{|c|c|c|c|c|c|c|}
		\hline
		i & $q_1$  & $q_2$ & $q_3$ & $q_4$ & $q_5$ & $q_6$ \\
		\hline
		1  &  1.108 & 2.597 &  2.062 & -2.896 &  2.690 &  0.498 \\
        2  &  0.900 & 0.953 &  0.858 & -2.601 & -0.169 & -0.334 \\
        3  &  0.312 & 1.050 &  0.713 & -1.941 & -0.679 & -1.167 \\
        4  &  3.129 & 0.259 &  0.688 & -1.546 &  2.241 & -1.836 \\
        5  & -3.050 & 2.077 &  1.912 & -1.436 &  0.824 &  1.165 \\
        6  & -0.393 & 2.850 &  1.922 & -1.186 & -1.762 &  1.817 \\
        7  & -2.230 & 0.461 &  0.861 & -0.526 &  2.983 & -0.318 \\
        8  & -2.226 & 2.191 &  2.003 & -0.521 &  0.150 &  0.721 \\
        9  & -2.035 & 0.543 &  1.080 &  0.248 &  2.691 &  0.500 \\
        10 & -2.240 & 2.188 &  2.283 &  0.538 & -0.168 & -0.332 \\
        11 & -2.827 & 2.090 &  2.428 &  1.198 & -0.677 & -1.165 \\
        12 & -0.013 & 2.881 &  2.452 &  1.593 &  2.240 & -1.837 \\
        13 &  0.089 & 1.063 &  1.229 &  1.703 &  0.826 &  1.165 \\
        14 &  2.749 & 0.290 &  1.219 &  1.953 & -1.764 &  1.817 \\
        15 &  0.909 & 2.680 &  2.280 &  2.613 &  2.983 & -0.320 \\
        16 &  0.913 & 0.949 &  1.138 &  2.618 &  0.151 &  0.722 \\
		\hline
	\end{tabular}
	\caption{16 inverse kinematic solutions for the pose
    $[-0.2406,  -0.1188,  0.5603,  2.6204,  1.1236,  0.4276]$ given in radians.}
	\label{tab:sol_iks}
\end{table}
\subsection{Validation of a robot trajectory}
For each initial posture, we will execute the trajectory and evaluate, at each point, the proximity to the robot's singularities and joint limits. To move the robot, we use the kinematic Jacobian matrix ${\bf J^{-1}}$, which relates the kinematic twist ${}^0\mathbb{V}_E$ (of the end effector expressed in the frame attached to the base) to the joint velocities vector $\dot{\textbf{q}}$ (see \cite{sciavicco2010robotics}):
\begin{equation}
    \dot{\textbf{q}} = \textbf{J}^{-1}(\textbf{q}) {}^0\mathbb{V}_E\text{ .}
\end{equation}
%
The $i^{\rm th}$ column of this Jacobian matrix for a revolute joint can be computed from the Direct Kinematic Model (DKM), with ${}^0\textbf{P}_{i}$ being the position vector of frame $i$ in the robot's base frame and the local vector $z_i$ expressed in the robot's base frame\footnote{The \lq hat' notation is $(3 \times 3)$ asymmetric tensor associated to ($3 \times 1$) mentioned vector. For any vectors $U$ and $V$ of $\mathbb{R}^3$, $\hat{U}$ is defined such that $\hat{U}V = U \times V$.}:
\begin{equation}
    {}^0\textbf{J}_{E_i}=\begin{bmatrix}{}^0\hat{z}_{i}({}^0\textbf{P}_{n}-{}^0\textbf{P}_{i})\\{}^\textbf{0}z_{i}\end{bmatrix}\text{ .}
\end{equation}
%
In our application, we are not concerned with the joint velocities but rather with their successive positions. We proceed iteratively by approximating the integral using the explicit Euler method. At each iteration, we move by a small pose $\textbf{dk}$ (which is a change in position and orientation of the end effector), i.e:
\begin{equation}
    \textbf{q}_{j+1} \simeq \textbf{q}_j + \textbf{J}^{-1}(\textbf{q}_j)\textbf{dk}\text{ .}
\end{equation}
%
The trajectory is represented by a set of points connected by segments. Along these segments, change in position and orientation are assumed to be uniform. The position increment step $dp$ is chosen to guarantee an acceptable accuracy of motion \cite{paul1981robot}. This implies $|\textbf{dk}_P|=dp$, where $\textbf{dk}_P$ denotes the position component of $\textbf{dk}$. The intensity of the orientation change $\textbf{dk}_\omega$ is defined as performing the orientation change over the same number of increments as the position along the segment. For a given segment, we can introduce $\textbf{p}_{p_{i+1}}-\textbf{p}_{p_i}$ the change of position between pose $p_i$ and $p_{i+1}$ and $\textbf{R}=\textbf{R}_{p_{i+1}}\textbf{R}_{p_i}^T$ the change of rotation. The number of simulation steps $N$ and the variations in $\textbf{dk}$ are given by:
\begin{equation*}
    \textbf{N} = \dfrac{|\textbf{p}_{p_{i+1}}-\textbf{p}_{p_i}|}{dp} \text{ , }
    \textbf{dk}_P = \dfrac{\textbf{p}_{p_{i+1}}-\textbf{p}_{p_i}}{N} \text{ , and }
    \textbf{dk}_{\omega} = \dfrac{\theta\textbf{u}}{N} \text{ , }
\end{equation*}
%
where $\theta$ and $\textbf{u}$ are the representation of the rotation matrix $\textbf{R}$ in the form of a rotation angle  and a rotation axis. It can be computed using Rodrigues formula \cite{liang2018efficient}:

\begin{equation}
\theta = \text{acos}\left(\dfrac{\text{trace}(\textbf{R})-1}{2}\right)
\text{, and } 
\textbf{u} = \dfrac{1}{2S_{\theta}}\begin{bmatrix}
            {r}_{32} - {r}_{23}\\
            {r}_{13} - {r}_{31} \\
            {r}_{21} - {r}_{12}\\
        \end{bmatrix}\text{ .}
\end{equation}

%
%
Along the trajectory, we will assess its feasibility by ensuring that none of the robot's joints exceeds the joint limits given in Table \ref{tab:limit_articulaire}, that the determinant of the Jacobian matrix does not become singular, which would result in the loss of mobility and very high joint speeds \cite{yoshikawa1985manipulability}, and that the robot does not self-collide (see the next section). Algorithm \ref{alg:play_traj} summarizes the procedure to simulate and plan a trajectory.

\begin{algorithm}
    \caption{Simulate a trajectory}
    \label{alg:play_traj}
    \begin{algorithmic}
        \Require ${}^W {\bf T}_{p}, {}^W {\bf T}_0, {}^6{\bf T}_E$
        \For{$i \in N_{points}$}
            \State ${}^0{\bf T}_{6}^{(i)} = {}^W{\bf T}_0^{-1}.{}^W{\bf T}_{p}^{(i)}.{}^6{\bf T}_E^{-1}$
        \EndFor

        \State $\text{postures} = \text{IKM}({}^0{\bf T}_{6}^{(0)})$

        \For{${\bf q}_0 \in \text{postures}$}
            \State ${\bf q}_j = {\bf q}_0$
            \For{$ i \in [0, N_{points}-1]$} 
                \State $[N_{incr}, dk] = f({}^0{\bf T}_{6}^{(p_i)}, {}^0{\bf T}_{6}^{(p_{i+1})}, dp)$
                \For{$ \_ \in [0,N_{incr}]$}
                    \State $\textbf{q}_{j+1} = \textbf{q}_j + \textbf{J}^{-1}(\textbf{q}_j)\textbf{dk}$
                    \State $d_{sing} = \det(\textbf{J}({\bf q}_{j+1}))$
                    \State $d_{lim} = \min({\rm dist}(\textbf{q}_{j+1}, \textbf{q}_{limit}))$
                    \State $\text{coll} = \text{robot\_self\_collision}(\textbf{q}_{j+1})$
                \EndFor
            \EndFor
        \EndFor
    \end{algorithmic}
\end{algorithm}

\subsection{Detection of robot self-collisions}

Along the trajectory, the robot may self-collide even if it remains within its joint limits, e.g. when the elbow is bent towards the base. It is important to determine these configurations, which will prevent the trajectory from being executed on the physical robot. Moreover, these calculations must be kept as simple as possible, to avoid a significant increase in computation times. 


We adopted a method based on the computation of the minimum distance between two segments. To do this, we implemented an algorithm proposed by \cite{lumelsky1985fast}, designed to reduce the number of calculations required. This algorithm proceeds as follows:

\begin{enumerate}
    \item It begins by extending the segments into straight lines and identifying the points that minimize the distance between them. This is done using the parametric representation of each line to compute the distance. The partial derivative is zero for the minimum distance.
    \item If these points do not lie on the segments, the first point is moved to the nearest extremity, and the nearest corresponding point on the other line is then computed.
    \item If the second point is not on the segment, it is shifted to its nearest extremity, and a new corresponding point is determined on the first line.
\end{enumerate}
%
%
Remark, in the case of parallel lines, select one end of the first segment and go directly to step 3. 
%
%

The developed method is based on the successive computation of the minimum distances between robot segments. First, the coordinates of the segment extremities are obtained using the DKM, which provides the positions of the joint centres and the end effector.
%
%
The algorithm then proceeds iteratively: for each segment, it computes the minimum distance to other segments that are not successive (and have not yet been evaluated). The robot's segments are modeled as capsules, and the radius of each capsule is subtracted to obtain the final minimum distance.
%
%
If one of these distances is smaller than or equal to zero, the robot is in collision. The whole process is summarized in the algorithm \ref{alg:collision}.

\begin{algorithm}
    \caption{Robot self-collisions}
    \label{alg:collision}
    \begin{algorithmic}
        \Require $\textbf{q}$
        \State $\textbf{P} = \text{DKM}(\textbf{q})$
        \For{$i \in [0, N-2]$}
        \For{$j \in [i+2, N]$}
            \State $\text{Seg}_i = (\textbf{P}_i, \textbf{P}_{i+1})$
            \State $\text{Seg}_j = (\textbf{P}_j, \textbf{P}_{j+1})$
            \State $\text{dist}_{min} = \text{min\_distance\_segments}(\text{Seg}_i, \text{Seg}_j)$
            \If{$\text{dist}_{min} \le R_{\text{caps}_i} + R_{\text{caps}_j}$}
                \State return True
            \EndIf
        \EndFor
    \EndFor
    \State return False
    \end{algorithmic}
\end{algorithm}

%
%
\section{Optimization of the Robot's Position}

Using the developed simulator, we can assess the feasibility of a trajectory for a given robot configuration, defined by the position and orientation of its base and TCP. This allows us to explore various configurations and give them a score.
%
%
In our application, the robot base is mounted on a mobile support, controlled by the user. Our goal is to assist him in selecting the optimal placement. As the positioning accuracy is approximate, we prioritize an approach that is robust to these inaccuracies rather than focusing solely on singularities and joint limits. Specifically, our aim is to identify the centre of the largest inscribed circle within the trajectory's feasibility region.
\subsection{Exploring configuration spaces with a PSO algorithm}

The search space is potentially vast, it is not conceivable to exhaustively evaluate all possible configurations in a production environment as the simulation time of a configuration is too high. To overcome this constraint, we have used a PSO algorithm\cite{eberhart1995new} to efficiently explore the search space. The aim is first to investigate the whole space, then to concentrate on areas of interest, i.e. those where the trajectory is feasible.
%
Particle swarm optimization is inspired by the collective behavior of birds in flight and schools of fish. This algorithm relies on a population of particles exploring the search space to identify an optimal solution. Each particle represents a candidate solution and is characterized by:

\begin{enumerate}
    \item An $\bf x_i$ position: corresponds to a potential solution to the problem. Note that the orientation angle $\psi$ (see (\ref{eq:wte}) is chosen so that the robot faces the workpiece.
    \item A velocity $\bf v_i$: controlling the particle's displacement in the search space.
    \item A personal history ${\bf p}_{best_i}$: the best position reached by the particle.
    \item A local history ${\bf l}_{best_i}$: the best position found by particles in its neighbourhood at iteration $i$. 
\end{enumerate}

At each iteration, the position and velocity of each particle are updated according to the following equations\footnote{The $\circ$ notation is the Hadamard product i.e. element-wise product of vectors or matrices \cite{el2015dynamic}.}:
\begin{eqnarray}
    {\bf v}_{i+1}&=&w {\bf v}_i + c_1{\bf r}_{i1}\circ({\bf p}_{best_i}-{\bf x}_i) \label{eqn:v_pso} \\
    &+&  c_2 {\bf r}_{i2} \circ({\bf l}_{best_i}-{\bf x}_i) + e {\bf r}_{ie}  \text{ , and }\nonumber \\ 
    {\bf x}_{i+1}&=&{\bf x}_i+{\bf v}_{i+1}\text{ , }
\end{eqnarray}
%
where $w$, $c_1$, $c_2$ and $e$ are inertia, cognition, social, and exploration coefficients respectively, while ${\bf r}_{i1}$, ${\bf r}_{i2}$ and ${\bf r}_{ie}$ are random vectors of $\mathbb{R}^n$ uniformly distributed in $[0,1]^n$. The subset of particles forming a neighborhood is defined by a structure called topology.

%
%
Topology defines the way particles communicate with each other to share information about the best solutions found. It has a significant influence on the convergence rate and exploration ability of the algorithm. Static topologies are not suitable for all optimization problems \cite{kennedy2002population}. Thus, we used a dynamic topology called DCluster \cite{el2015dynamic}, which combines two existing approaches: Four-clusters and Fitness. 
%
%
The DCluster topology is based on a dynamic segmentation of the swarm into several sub-swarms of equal size, called clusters. At each iteration, the topology is built in four steps:

\begin{enumerate}
    \item Evaluate and sort particles according to their score for the current position, in ascending order.
    \item Split the sorted swarm into several clusters of equal size, where the last cluster contains the best particles.
    \item Fully connect particles in the same cluster, to create strongly connected subgraphs.
    \item Create a central cluster, composed of the worst-performing particles, and connect each of them to the worst particle in a higher cluster.
\end{enumerate}
%
This topology implies a specific swarm size given by the following equation:
\begin{equation}
    S = N \times (N+1) \text{ , }
\end{equation}
%
where $N$ is the number of particles per cluster and $S$ the total number of particles.
%
This structure promotes efficient exploration of the search space by slowing down the propagation of information between clusters, thus reducing the risk of premature convergence towards a local optimum.

%
%
In our specific problem, the function to minimise corresponds to the percentage of the reference trajectory that could not be completed. This percentage is evaluated in terms of the number of segments covered. In contrast to conventional PSO applications, the minimum value is known, and the optimal solution is not necessarily reduced to a single point, but rather to a set of surfaces.
%
%
Consequently, the objective is not to converge on a single solution, but to explore the whole space before focusing on the boundaries of the feasible zones. To encourage this exploration, we have introduced an exploratory factor $e$, see (\ref{eqn:v_pso}).
%
%
Moreover, because we know in advance the space to be explored, we have adopted a different approach to conventional PSO algorithms. The particles are initially distributed uniformly along the boundaries of the space, forming a circle. The initial velocity is directed towards the center of the trajectory. This strategy aims to maximize coverage of the search space and to avoid missing any zones of interest. The algorithm \ref{alg:pso} synthesizes the PSO method used to explore the parameter space. 

\begin{algorithm}
    \caption{Particle Swarm Optimisation Algorithm}
    \label{alg:pso}
    \begin{algorithmic}
    
    \State Initialize particle positions ${\bf x}_0$ uniformly on the search space boundary 
    \State Set corresponding velocities ${\bf v}_0$ towards the trajectory center.
    \State Evaluate $f({\bf x}_0)$   
    \For{each iteration $i$}
        \State Sort particles by fitness $f({\bf x}_i)$ in ascending order and divide them into $N$ clusters of equal size
        \State Fully connect particles within each cluster
        \State Connect each particle from the worst cluster to the worst particle of a superior cluster
        \For{each particle}
            \State ${\bf l}_{best_{i}} = {\bf p}_{best_{i,k}}$ where $ P_k = {\rm argmin}\left( f({\bf p}_{best_{i,j}}), P_j \in {\rm neighborhood}\right)$
            \State ${\bf v}_{i+1} = w {\bf v}_i + c_1{\bf r}_{i1} \circ ({\bf p}_{best_i} - {\bf x}_i) + c_2 {\bf r}_{i2} \circ ({\bf l}_{best_i} - x_i) + e{\bf r}_{ie}$
            \State ${\bf x}_{i+1} = {\bf x}_i + {\bf v}_{i+1}$
        \EndFor
        \For{each particle}
            \State Evaluate $f({\bf x}_{i})$
            \If{$f({\bf x}_i) \le f({\bf p}_{best_i})$}
                \State ${\bf p}_{best_i} = {\bf x}_i$
            \EndIf
        \EndFor
    \EndFor
    \end{algorithmic}
\end{algorithm}
\subsection{Clustering and determination of the largest inscribed circle}
The PSO algorithm allows us to efficiently explore the search space and identify areas of trajectory feasibility. We can then analyse these areas to determine the center of the largest inscribed circle. After filtering the point cloud to retain only the relevant points, we draw the polytopes delimiting these regions.
%
%
To do this, we use the $\alpha$-shape algorithm \cite{edelsbrunner1983shape}, which extracts the envelope of a 2D point cloud.  Unlike the classic convex hull method, $\alpha$-shape offers more flexibility by avoiding convex over-approximations and by adapting better to the real geometry of the data. This can be done by controlling the level of concavity with a parameter $\alpha$. The method also identifies several clusters of points, representing different regions accessible to the robot, as well as any holes in the point cloud, which may represent under-explored areas or real regions of infeasibility.

%
%
The algorithm proposed in this paper is based on the following principle: a sphere of radius $\alpha$ explores the space containing the cloud points. Two points are connected by an edge in the $\alpha$-shape if there is a disk of radius $\alpha$ containing them on its boundary without including any other points inside. The set of edges then forms the polytope delimiting the point cloud.
A method based on Delaunay triangulation \cite{delaunay1934sphere} can be:
\begin{enumerate}
    \item Construct the Delaunay triangulation for all points $S$.
    \item For each simplex, calculate its circumscribed radius $r$.
    \item Include the simplex in the $\alpha$-complex if $r\le \alpha$.
    \item The resulting $\alpha$-complex boundary forms the $\alpha$-shape.
\end{enumerate}
\bigskip
Once the polytopes have been determined for each cluster, we can then proceed to determine the largest circle inscribed in each polytope. To do this, we use an approach based on Voronoi diagrams \cite{beyhan2020algorithm}:
\begin{enumerate}
    \item Calculate the Voronoi diagram of the polygon.
    \item Identify diagram nodes inside the polygon.
    \item For each internal node, measure the minimum distance to the polygon's edges.
    \item Select the node with the maximum distance, defining the center of the largest inscribed circle, with this distance corresponding to the radius.
\end{enumerate} 
In order to obtain a Voronoi diagram representing the whole polygon and not just its vertices, we discretize the polygon edges. Note that this method can also be applied to non-convex polygons with holes. Clusters with a maximum radius of less than 0.05 m are not taken into account, as the margins are too small to be placed by a user.
\section{Experimentations}
%
This section presents the results obtained for an industrial plasma cutting application. Specifically, it focuses on a trajectory resulting from the intersection of two cylinders. This application can be found, for example, in the fields of shipbuilding, boiler making, or metal infrastructures, for the assembly of piping networks or tubular structures.
%
%
The objective is to cut the larger diameter tube so that the second tube can be fitted into it, ensuring a precise fit for subsequent welding operations. The trajectory is shaped like a horse saddle and will be referred to as ``saddle'' in the remainder of this document.
%
%
In addition to this base trajectory, a first lead-in is used to allow time to pierce the thickness of the tube and reach the circle tangent, thus ensuring a clean cut and avoiding geometrical imperfections related to the cutting process.
%
%
The path positions depend on the diameters of the two tubes and the angle of inclination between them. The path orientation, on the other hand, depend on a $\alpha$ bevel angle (determined according to the requirements of the welding process) and the presence or absence of a camera in the cutting axis.

\begin{figure}[!ht]
	\centering
	\begin{subfigure}{0.17\textwidth}
		\includegraphics[width=\textwidth]{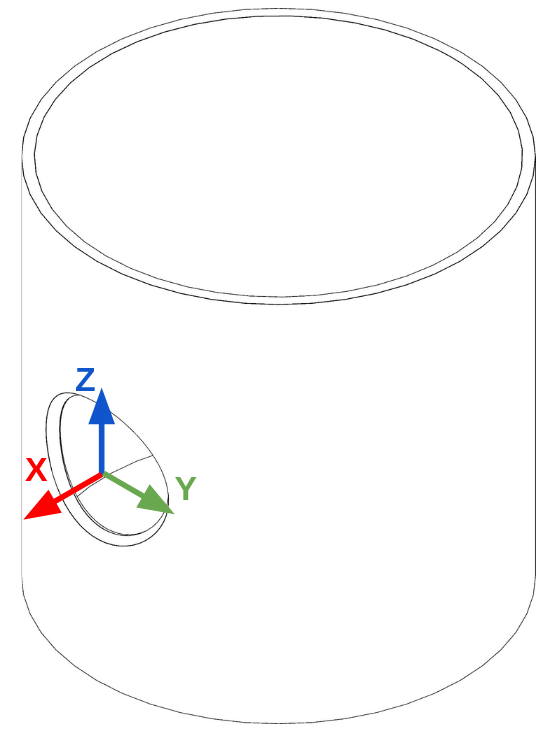}
		\caption{Workpiece}
	\end{subfigure}
	\begin{subfigure}{0.22\textwidth}
		\includegraphics[width=\textwidth]{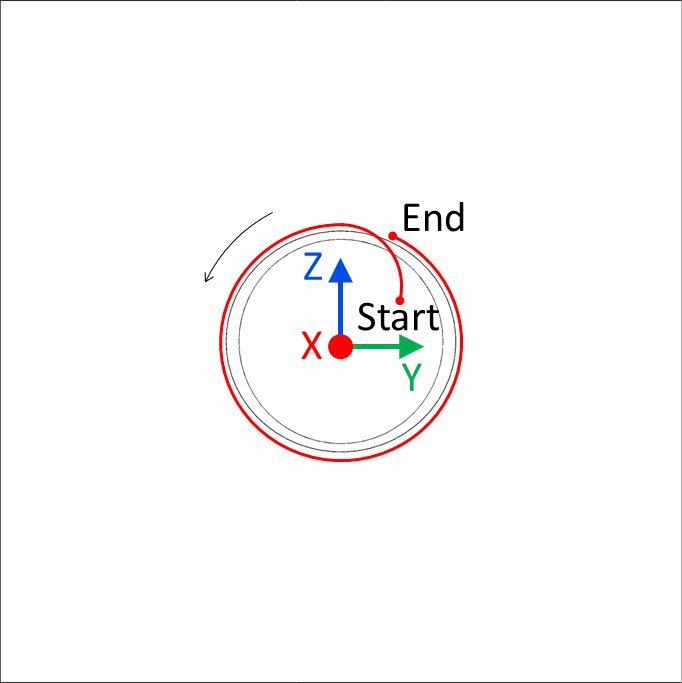}
		\caption{Trajectory}
	\end{subfigure}
	\caption{Workpiece and its associated saddle-shaped trajectory.}
	\label{fig:traj_saddle}
\end{figure}

\begin{figure}[!ht]
	\centering
	\begin{subfigure}{0.42\textwidth}
		\includegraphics[width=\textwidth]{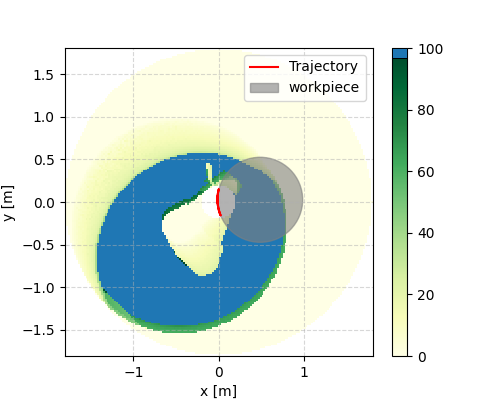}
		\caption{$\beta=15^o$}
	\end{subfigure}
	\begin{subfigure}{0.42\textwidth}
		\includegraphics[width=\textwidth]{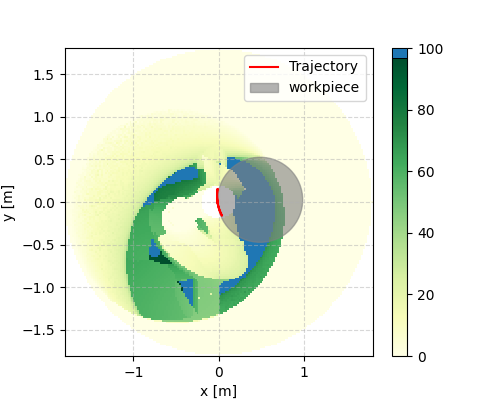}
		\caption{$\beta=45^o$}
	\end{subfigure}
	\begin{subfigure}{0.42\textwidth}
		\includegraphics[width=\textwidth]{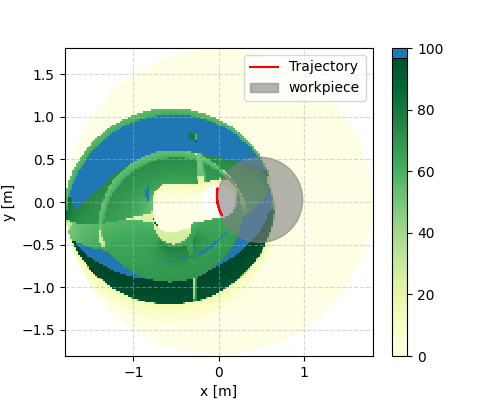}
		\caption{$\beta=15^o$ and camera in axis}
	\end{subfigure}
	\caption{Percentage of feasible trajectory for different saddle-shaped trajectories.}
	\label{fig:perc_traj}
\end{figure}

%
%
In this section, we consider an example where the diameter of the larger tube is 1 m and the smaller tube is 0.3m, the larger tube is vertical and the smaller is perpendicular. The TCP frame ${}^6{\bf T}_E$ used corresponds to the pose [0.038, 0, 0.409, $\pi/4$,0,-$\pi/2$]. Figure~\ref{fig:traj_saddle} shows the final workpiece and a visualization of the trajectory to be performed. 
%
%
For a given tube configuration i.e., the radius of the cylinders and their relative orientation. The choice of angles can significantly influence the trajectory's feasibility zones. As an example, Figure~\ref{fig:perc_traj} shows three different configurations of chamfer angles $\beta$ and camera used: (a) 15° chamfer, without camera in axis; (b) 45° chamfer, without camera in axis, and (c) 15° chamfer, with camera in axis. This last property allows the user to control the execution of the trajectory and adapt it in terms of position and speed in real-time.
%
%
Each point in Figure~\ref{fig:perc_traj} represents the percentage of the trajectory that can be achieved before reaching a non-feasibility factor, such as a singularity, a joint limit, or a collision, for the best-associated posture. The blue dots represent the points where the entire trajectory is achievable. Figure~\ref{fig:perc_traj} shows the different feasibility zones in the x-y plane for a height $z=-0.12$m from the center of the trajectory (center of the circle). The trajectory is projected onto the plane in red and the cylinder is represented by the grey circle. Note that the robot cannot be placed in this area. The results obtained highlight the difficulties the user may face when placing the robot. In fact, feasibility zones are highly dependent on the trajectory to be achieved, and a simple modification of a parameter can lead to a drastic change in their shape and size.
%
%
The computation times required to exhaustively simulate all combinations are not compatible with online programming. The approach developed in this article, based on a PSO algorithm, enables us to efficiently explore the entire search space and concentrate on the boundaries of feasibility zones.
%

%
%

\begin{figure}[htbp]
	\centering
        \begin{subfigure}{0.45\textwidth}
		\includegraphics[width=\textwidth]{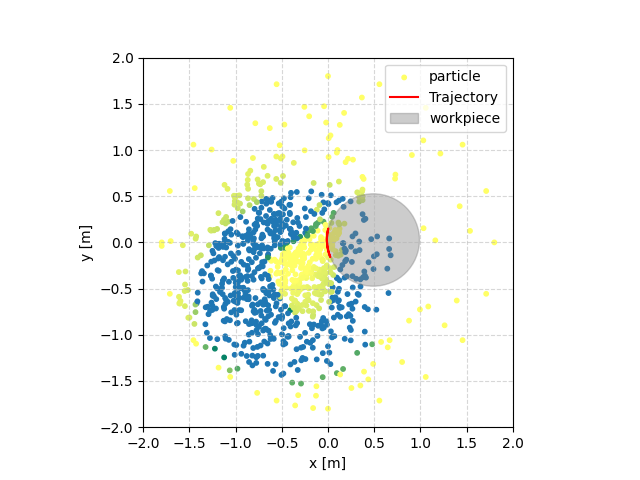}
		\caption{$\beta=15^o$}
	\end{subfigure}
	\begin{subfigure}{0.45\textwidth}
		\includegraphics[width=\textwidth]{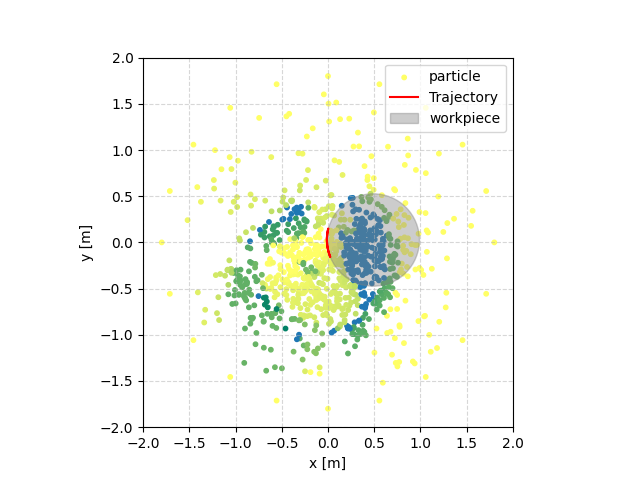}
		\caption{$\beta=45^o$}
	\end{subfigure}
	\begin{subfigure}{0.45\textwidth}
		\includegraphics[width=\textwidth]{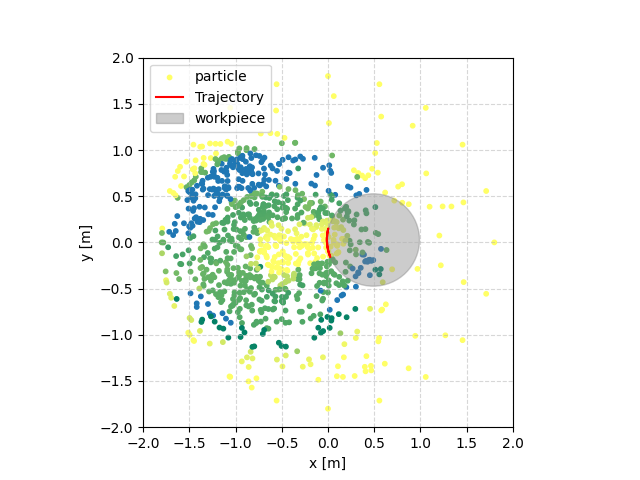}
		\caption{$\beta=15^o$ and camera in axis}
	\end{subfigure}
	\caption{Exploration with the PSO algorithm on the previous examples: gradient from green to yellow indicates the score in ascending order when blue is the minimal score.}
	\label{fig:pso}
\end{figure}

We have applied this algorithm to the examples presented above. Figure~\ref{fig:pso} illustrates the set of combinations explored by the PSO algorithm for previous examples. The blue dots represent configurations where the trajectory was entirely feasible, while a gradient from green to yellow indicates the score of the cost function in ascending order.
%
In these examples, the parameters of the PSO algorithm used are: $N=4$, $iter=50$, $w=0.8$, $c_1=0.35$, $c_2=0.15$, and $e=0.2$.  For comparison, simulation times are less than 3 minutes (1000 configurations tested), while they can reach 1 hour for exhaustive exploration ($\simeq 20,000 $ configurations tested). In each simulation, the particles first explored the entire space before focusing on the areas of interest.

%
%

\begin{figure}[!ht]
	\centering
        \begin{subfigure}{0.46\textwidth}
		\includegraphics[width=\textwidth]{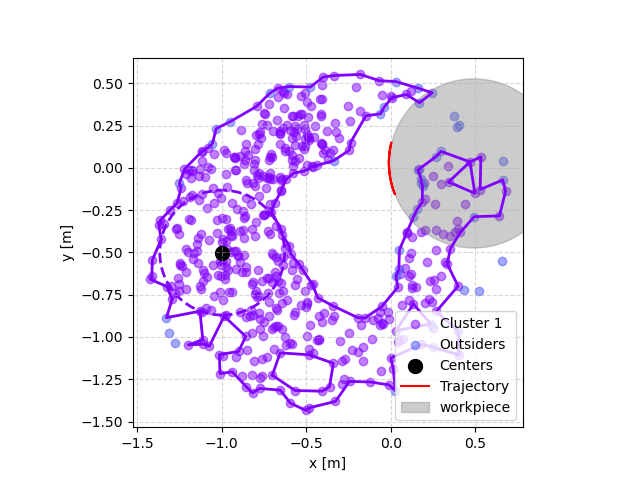}
		\caption{$\beta=15^o$}
	\end{subfigure}
	\begin{subfigure}{0.46\textwidth}
		\includegraphics[width=\textwidth]{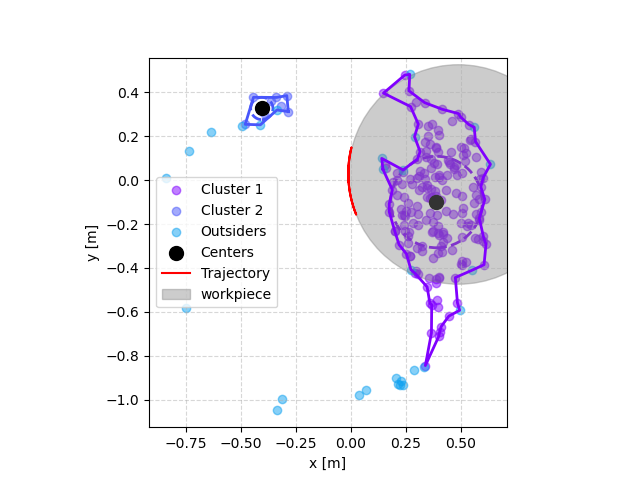}
		\caption{$\beta=45^o$}
	\end{subfigure}
	\begin{subfigure}{0.46\textwidth}
		\includegraphics[width=\textwidth]{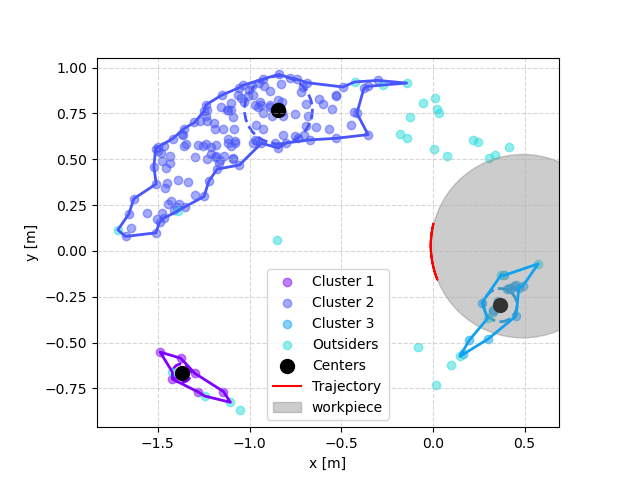}
		\caption{$\beta=15^o$ and camera in axis}
	\end{subfigure}
	\caption{Clusters calculated with the $\alpha$-shape algorithm and their largest inscribed circles on the previous examples.}
	\label{fig:alpha_shape}
\end{figure}

From the positions explored, retaining only those for which the trajectory is feasible, we can construct the $\alpha$-shape and compute the largest inscribed circles. Figure~\ref{fig:alpha_shape} shows the results obtained for the previous configurations. For these examples, we used a parameter $\alpha = 0.05$m. In each case, the main cluster was correctly identified, and the largest inscribed circle corresponds closely to the one that would have been obtained using exhaustive discretization (see Figure \ref{fig:perc_traj}). The center and radius of this circle provide valuable assistance to the user in the placement of the robot, indicating both the optimum position and the tolerance margin available. For example, in case (c), the algorithm returns the point x=-1.37m, y=-0.66m and with a margin of 0.18m. The positions returned by the algorithm were tested on the physical Fanuc robot. In each case, the trajectory was achieved without any problems. Note that for the second example, the second cluster was used because the first one was behind the workpiece.

\begin{figure}[!ht]
	\centering
	\includegraphics[width=0.45\textwidth]{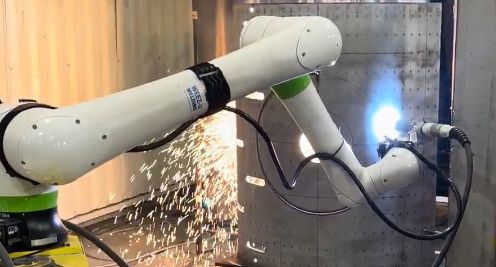}
	\caption{Optimizing robot positioning for plasma cutting is an industrial application.}
	\label{fig:welding}
\end{figure}
\section{Conclusions}
In this study, we have developed an innovative approach to optimizing the placement of robotic arms, using the Fanuc CRX10iA/L robot as a case study. The trajectory simulation incorporates motion constraints, singularities, joint limits, and robot self-collision detection to ensure feasibility and efficiency.
%
By integrating kinematic simulation, a particle swarm optimization algorithm, and feasibility zone identification using the $\alpha$-shape algorithm, we have developed an efficient and robust method for optimizing robot placement.

%
%


Experimental results have convincingly demonstrated that our approach not only effectively identifies feasibility zones but also provides a robust criterion for handling initial placement uncertainties. Its industrial application to a plasma cutting task has confirmed the method's relevance, significantly reducing calibration time and improving the reproducibility of robot-executed trajectories. Moreover, cutting under real shop conditions was successfully executed using the Weez-U welding mobile base, as illustrated in Figure~\ref{fig:welding}.

%
%
Future directions for this study include integrating the workspace into the simulation to account for complex parts a robot may need to manipulate. Additionally, extending this approach to other optimization parameters will assist robotic integrators in application design, particularly in selecting a robust TCP. Finally, optimizing computation times by translating the Python code into a lower-level language and parallelizing simulations would enable the exploration of more parameter dimensions without excessive computational costs. This would also allow fast optimal robot positioning, enhancing its usability in online programming.
\section*{Acknowledgement}
This research was supported by ANRT CIFRE grant n°2023 /1565 which funded the first author's doctoral studies. 
\selectlanguage{english}
\bigskip
\bibliographystyle{unsrt}
\bibliography{asme_reference}
\end{document}